\documentclass[journal]{IEEEtran}
\usepackage{subfig}
\usepackage{multirow}
\usepackage{epsfig}
\usepackage{graphicx}
\usepackage{color}
\usepackage{amsmath}
\usepackage{amssymb}

\begin{document}
	%
	\title{PSC-Net: Learning Part Spatial Co-occurrence for Occluded Pedestrian Detection}

	
	\author{
		\IEEEauthorblockN{Jin Xie\IEEEauthorrefmark{1},
			Yanwei Pang\IEEEauthorrefmark{1},
			Hisham Cholakkal\IEEEauthorrefmark{2}, 
			Rao Muhammad Anwer\IEEEauthorrefmark{2}, 
			Fahad Shahbaz Khan\IEEEauthorrefmark{2}, and
			Ling Shao\IEEEauthorrefmark{2}}\\
		\IEEEauthorblockA{\IEEEauthorrefmark{1}Tianjin University}
		\IEEEauthorblockA{\IEEEauthorrefmark{2}Inception Institute of Artificial Intelligence}
	}

	

	\IEEEtitleabstractindextext{%
		\begin{abstract}
Detecting pedestrians, especially under heavy occlusions, is a challenging computer vision problem with numerous real-world applications. This paper introduces a novel approach, termed as PSC-Net, for occluded pedestrian detection. The proposed PSC-Net contains a dedicated module that is designed to explicitly capture both inter and intra-part co-occurrence information of different pedestrian body parts through a Graph Convolutional Network (GCN). Both inter and intra-part co-occurrence information contribute towards improving the feature representation for handling varying level of occlusions, ranging from partial to severe occlusions. Our PSC-Net exploits the topological structure of pedestrian and does not require part-based annotations or additional visible bounding-box (VBB) information to learn part spatial co-occurrence.   
Comprehensive experiments are performed on two challenging datasets: CityPersons and Caltech datasets. The proposed PSC-Net achieves state-of-the-art detection performance on both. On the heavy occluded (\textbf{HO}) set of CityPerosns test set, our PSC-Net obtains an absolute gain of 4.0\% in terms of log-average miss rate over the state-of-the-art~\cite{MGAN_2019_ICCV} with same backbone, input scale and without using additional VBB supervision. Further, PSC-Net improves the state-of-the-art~\cite{Chunluan_2019_ICCV} from 37.9 to 34.8 in terms of log-average miss rate on Caltech (\textbf{HO}) test set.
		\end{abstract}
	}
	\maketitle
	\IEEEdisplaynontitleabstractindextext
	\IEEEpeerreviewmaketitle

	\section{Introduction}
	\IEEEPARstart{P}{edestrian} detection is a challenging problem in computer vision with various real-application applications, \emph{e.g.}, robotics, autonomous driving and visual surveillance. Recent years have witnessed significant progress in the field of pedestrian detection, mainly due to the advances in deep convolutional neural networks (CNNs). Modern pedestrian detection methods can be generally divided into single-stage \cite{WeiLiuECCV18, JunhyugNohCVPR18, Brazil_2019_CVPR} and two-stage \cite{Songtao_2019_CVPR, Chunluan_2018_ECCV, Chunluan_2019_ICCV, MGAN_2019_ICCV, ShanshanCVPR18, Zhang_2018_ECCV, GarrickBrazilICCV17, XinlongWangCVPR18, JiayuanMaoCVPR17}. Single-stage pedestrian detectors typically work by directly regressing  
	the default anchors into pedestrian detection boxes. Different to single-stage pedestrian detectors, two-stage methods first produce a set of candidate pedestrian proposals which is followed by classification and regression of these pedestrian proposals. Most existing two-stage pedestrian detectors \cite{Zhang_2018_ECCV, MGAN_2019_ICCV, Songtao_2019_CVPR, ShanshanCVPR18, XinlongWangCVPR18, JiayuanMaoCVPR17} are based on the popular Faster R-CNN detection framework~\cite{fasterrcnn_2015_nips} that is adapted from generic object detection. Existing pedestrian detectors typically assume entirely visible pedestrians when trained using full body pedestrian annotations.
	
	While promising results have been achieved by existing pedestrian detectors on standard non-occluded pedestrians, their performance on heavily occluded pedestrians is far from satisfactory. This is evident from the fact that the best reported performance~\cite{MGAN_2019_ICCV} on the reasonable (\textbf{R}) set (where visibility ratio is larger than 65\%) of CityPersons test set~\cite{citypersons_2017_zhang} is 9.3 (log-average miss rate) whereas it is 41.0 on the heavy occluded (\textbf{HO}) set (where visibility ratio ranges from 20\% to 65\%) of the same dataset. Handling pedestrian occlusion is an open problem in computer vision and present a great challenge for detecting pedestrians 
	in real-world applications due to its frequent occurrence. Therefore, a pedestrian detector is desired to be accurate with respect to varying level of occlusions, ranging from  reasonably occluded to severely occluded pedestrians.  
	
	A variety of occluded pedestrian detection approaches exist in literature. A common strategy to address occlusion is based on learning and integrating a set of part detectors \cite{YonglongTianICCV15, ChunluanZhouICCV17, WanliOuyangICCV13,MarkusMathiasICCV13, Ouyang_2013_CVPR, 2004_ECCV_part_label}.~Earlier part-based pedestrian detection approaches utilize body part annotations \cite{2004_ECCV_part_label, 2001_pami_part_label}. Recently, the deployment of pre-trained part models have been investigated \cite{ShanshanCVPR18} to exploit part correlation, typically relying on part detection scores corresponding to the visible regions of the pedestrian. Other methods \cite{YonglongTianICCV15, ChunluanZhouICCV17, WanliOuyangICCV13,MarkusMathiasICCV13, Chunluan_2016_ACCV_Part} utilize the bounding-box of the pedestrian and train a large number of independently learned part detectors.~Alternatively, the topological structure of the pedestrian has also been exploited \cite{Zhang_2018_ECCV} to avoid the reliance on body part annotations, leading to promising detection performance. However, it predominantly relies on the detection scores of parts to highlight visible regions of the pedestrian and do not consider spatial co-occurrence relations among different body parts \emph{and} between  different sub-regions (\emph{e.g.}, eyes  and  ears of a head region) within a body part. The part spatial co-occurrence information is expected to enrich the feature representation by exploiting the information about the spatially adjacent parts. Knowledge about the typical configuration of objects (\emph{e.g.}, humans) in a scene and its impact on recognition performance has been extensively studied previously in psychology and computer vision~\cite{Biederman_perception, Moshe_Bar_perception,Galleguillos_CVPR}. To the best of our knowledge, modern two-stage CNN-based pedestrian detectors do not explicitly encode the part spatial co-occurrence information. In this work, we introduce a data driven approach that goes beyond part detection scores and explicitly integrates part spatial co-occurrence information among different body parts \emph{and} between  different sub-regions within a body part to handle occlusions.
	
	\noindent \textbf{Contributions:} We propose a two-stage approach, termed as PSC-Net,~to address the problem of occluded pedestrian detection.~Our main contribution is the introduction of a novel part spatial co-occurrence (PSC) module designed to explicitly capture both inter and intra-part co-occurrence information of different pedestrian body-parts through a Graph Convolutional Network (GCN)~\cite{kipf_2017_semi_gcn}. Our PSC module only uses the standard full body supervision and neither requires part-based annotations nor rely on additional visible bounding-box (VBB) information.~To the best of our knowledge, we are the first to propose an approach based on GCN to capture both inter and intra-part spatial co-occurrence for occluded pedestrian detection.
	
\begin{figure}
		\centering
		\resizebox{1.0\linewidth}{!}{
			\begin{tabular}{c}
				\includegraphics[width=1.0\linewidth]{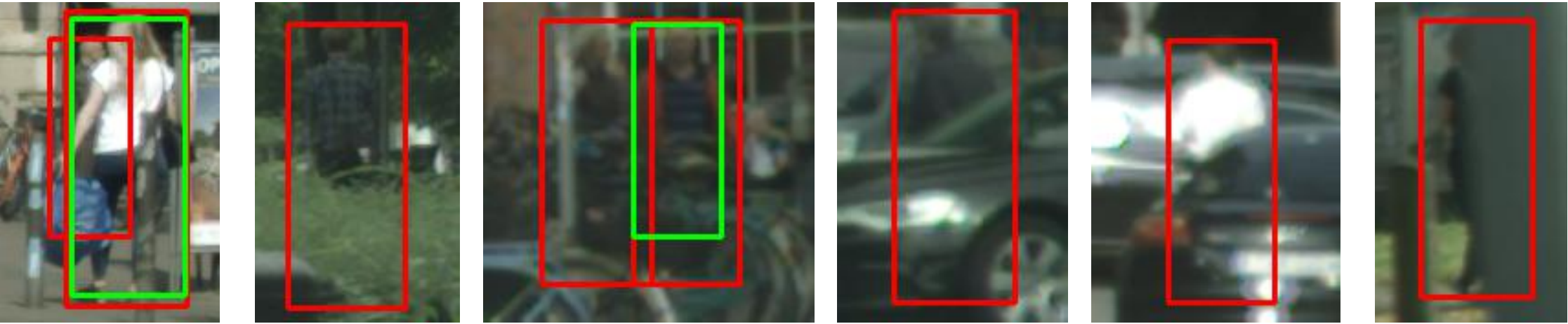}  \\
				(a) MGAN\cite{MGAN_2019_ICCV}  \\ 
				\includegraphics[width=1.0\linewidth]{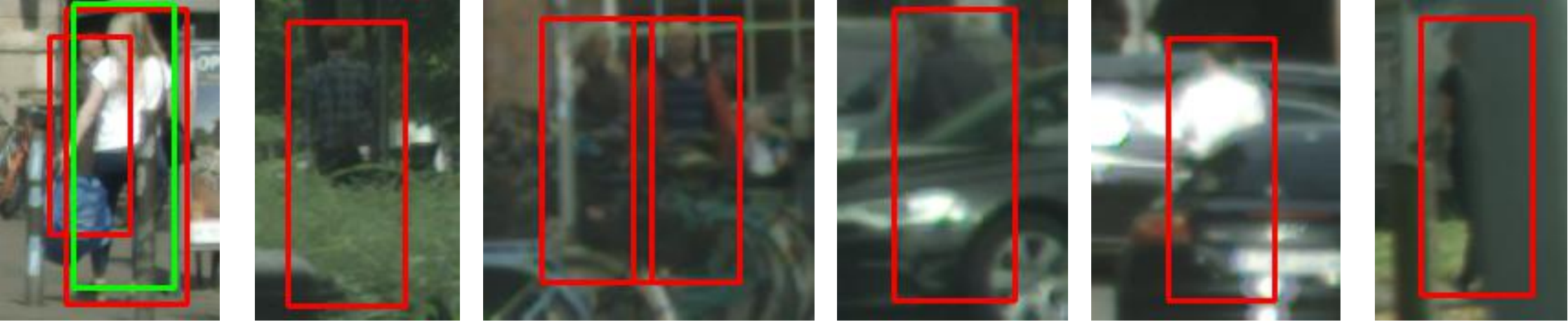}  \\
				(b) Baseline  \\ 
				\includegraphics[width=1.0\linewidth]{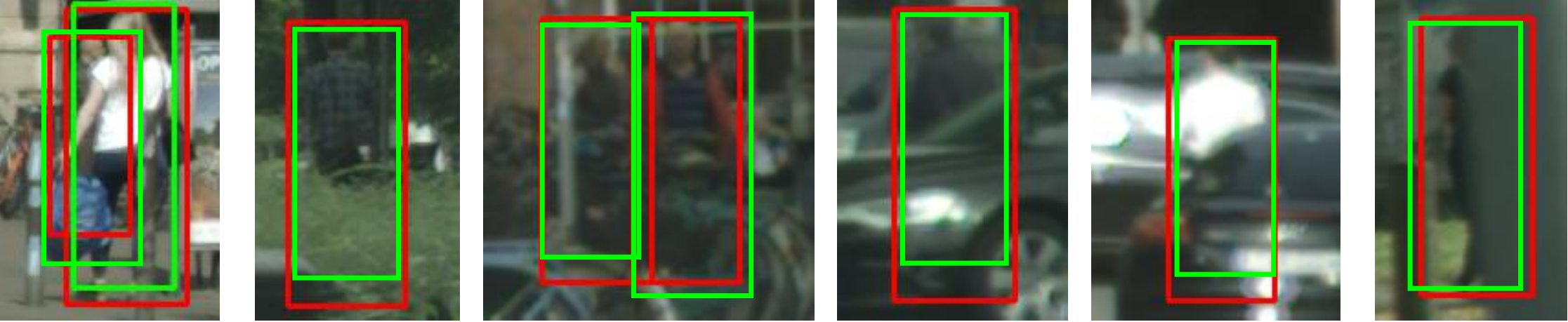}  \\
				(c) PSC-Net  \\          
		\end{tabular}}
		\caption{Qualitative detection examples using (a) the state-of-the-art MGAN~\cite{MGAN_2019_ICCV}, (b) our baseline and (c) our PSC-Net on CityPersons val. images. In the examples, red boxes denote the
			ground-truth and detector predictions are indicated by green boxes. The detected regions are cropped from the corresponding images for improved visualization. Note that all detection results are obtained using the same false positive per image (FPPI) criterion. Our PSC-Net  accurately detects pedestrians with varying level of occlusions.}  
		\label{fig:my_label_intro}
	\end{figure}

	We validate our PSC-Net by performing comprehensive experiments on two standard pedestrian detection datasets: CityPersons~\cite{citypersons_2017_zhang} and Caltech~\cite{Dollar_2012_PAMI}.~Our results clearly demonstrate
	that the proposed PSC-Net provides a significant improvement over the baseline.~On the heavy occluded (\textbf{HO}) set of CityPersons, our PSC-Net achieves an absolute gain of 5.5\% in terms of log-average miss rate, compared to the baseline. Further, PSC-Net sets a new state-of-the-art on both datasets. On the CityPerson (\textbf{HO}) test set, PSC-Net achieves a log-average miss rate of 37.4 with a significant gain of 3.6\% over the best reported results in literature~\cite{MGAN_2019_ICCV}. Fig.~\ref{fig:my_label_intro} shows example detections  with varying level of occlusions (partial to severe). Our PSC-Net is able to accurately detect pedestrians, compared to both the baseline and the state-of-the-art MGAN~\cite{MGAN_2019_ICCV}. 
	
	\begin{figure*}[t]
	\includegraphics[width=\textwidth]{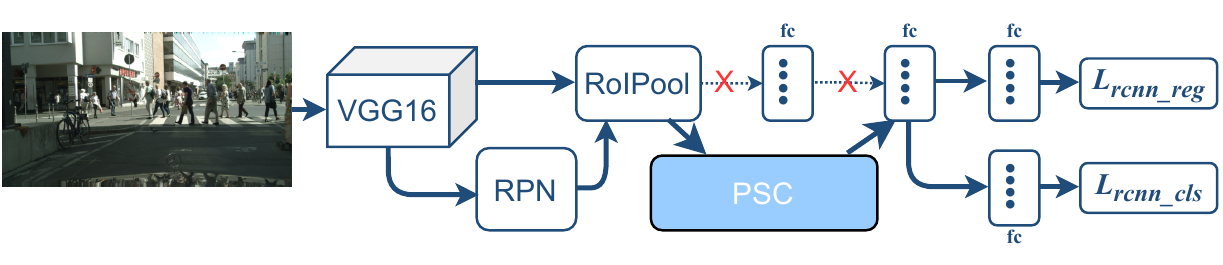}
	\caption{Overall network architecture of our PSC-Net. It consists of a pedestrian detection (PD) branch and a part spatial co-occurrence (PSC) module. In contrast to the baseline standard PD branch where the RoI features are used for box regression and classification (\textcolor{red}{X}), the RoI features in our PSC-Net are fed in to the proposed PSC module to integrate both intra and inter part spatial co-occurrence information.  The resulting enriched features are then deployed for final bounding-box regression and classification.}  
	\label{fig:overall_network}
\end{figure*}
	
	\section{Related Work}
	Convolutional neural networks (CNNs) have significantly advanced the state-of-the-art in numerous computer vision applications, such as image classification~\cite{He_2016_CVPR,simonyan2014vgg,Shao17INF,finegrained_AAAI2020}, object detection \cite{fasterrcnn_2015_nips,Pang_2019_CVPR_EFIP, Wang_2019_ICCV,Nie_2019_ICCV,liu2016ssd}, object counting \cite{zhang2015cross,cholakkal2019towards,Cholakkal_2019_CVPR,li2018csrnet}, image retrieval \cite{RadenovicECCV16,RadenovicPAMI19,LiuCYB19,HaofengTIP18}, action recognition \cite{SimonyanNIPS14,KhanTIP15,LiLiuTIP18,Narayan_2019_ICCV}, and pedestrian detection \cite{ShanshanCVPR18, MGAN_2019_ICCV,Zhang_2018_ECCV, XinlongWangCVPR18}. State-of-the-art pedestrian detection methods can generally be divided into single-stage and two-stage methods. Next, we present a brief overview of two-stage pedestrian detection methods. 
	
	\noindent \textbf{Two-stage Deep Pedestrian Detection:} In recent years, two-stage pedestrian detection approaches~\cite{citypersons_2017_zhang, MGAN_2019_ICCV, Chunluan_2018_ECCV, Chunluan_2019_ICCV, ShanshanCVPR18, Songtao_2019_CVPR, Zhang_2018_ECCV, XinlongWangCVPR18} have shown superior performance on standard pedestrian benchmarks. Generally, in two-stage pedestrian detectors, a set of candidate pedestrian proposals is first generated. Then, these candidate object proposals are classified and regressed. Zhang \emph{et al.},~\cite{citypersons_2017_zhang} propose key adaptations in the popular Faster R-CNN~\cite{fasterrcnn_2015_nips} for pedestrian detection. The work of~\cite{XinlongWangCVPR18} propose an approach based on a bounding-box regression loss designed for crowded scenes. The work of~\cite{ShanshanCVPR18} investigate several attention strategies, \emph{e.g.}, channel, part and visible bounding-box, for pedestrian detection. The work of~\cite{ZhaoweiCaiECCV16} introduce a multi-scale pedestrian detection approach with layers having receptive fields similar to object scales. Zhang \emph{et al.},~\cite{Zhang_2018_ECCV} propose a loss formulation that enforces candidate proposals to be close to the corresponding objects and integrates structural information with visibility predictions. The work of~\cite{Brazil_2019_CVPR} propose a multi-phase autoregressive pedestrian
	detection approach that utilizes a stackable decoder-encoder module with convolutional re-sampling layers. In~\cite{Songtao_2019_CVPR}, an adaptive NMS strategy is introduced that applies a dynamic suppression threshold to an instance. 
	
	\noindent \textbf{Towards Occluded Pedestrian Detection:} The problem of occluded pedestrian detection is well studied in literature~\cite{YonglongTianICCV15,ChunluanZhouICCV17,Chunluan_2018_ECCV,ShanshanCVPR18,XinlongWangCVPR18,Chunluan_2019_ICCV,Zhang_2018_ECCV,MGAN_2019_ICCV}. Some of these pedestrian detection approaches~\cite{YonglongTianICCV15,ChunluanZhouICCV17} handle occlusion by exploiting part-based information where typically a set of body part detectors are learned.~Each part is designated to handle a specific type (pattern) of
	occlusion.~Other approaches~\cite{XinlongWangCVPR18,Zhang_2018_ECCV,Songtao_2019_CVPR}  investigate novel loss formulations for detector training to improve pedestrian detection in crowded scenes under heavy occlusions.~Multiple pedestrian detectors  stacked in  a series are also investigated~\cite{WeiLiuECCV18, Brazil_2019_CVPR} to improve the detection performance.
	
	Most recent approaches~\cite{ShanshanCVPR18,Chunluan_2018_ECCV,Zhang_2018_ECCV,Chunluan_2019_ICCV,MGAN_2019_ICCV} investigate the problem of occluded pedestrian detection by utilizing additional visible bounding-box (VBB) annotations together with the standard full body information. Zhang \emph{et al.},~\cite{ShanshanCVPR18} use VBB along with a pre-trained body part prediction model to  tackle occluded pedestrian detection. The work of~\cite{Chunluan_2018_ECCV} demonstrate that an additional task of visible-region bounding-box prediction can improve the full body pedestrian detection. The work of \cite{Zhang_2018_ECCV} propose a novel loss that improves the localization, and  a part occlusion-aware region of interest pooling, to integrate structure information with visibility predictions. Zhou \emph{et al.},~\cite{Chunluan_2019_ICCV} propose a discriminative feature transformation module that projects the features in to a feature space, where the distance between occluded and non-occluded pedestrians is less. Such a transformation improves the robustness of the pedestrian detector during occlusion. In their approach, the visible bounding-box (VBB) is used to identify the occluded pedestrian. Recently, MGAN~\cite{MGAN_2019_ICCV} propose a mask-guided attention network,  using VBB annotation, to emphasize the visible regions while  suppressing the occluded regions, leading to state-of-the-art results on standard benchmarks.  
	
	\noindent \textbf{Our Approach:} Contrary to above mentioned recent approaches that rely on additional visible bounding-box (VBB) annotations, our proposed PSC-Net only requires the standard full body supervision to handle occluded pedestrian detection. The focus of our design is the introduction of a part spatial co-occurrence (PSC) module that explicitly captures both inter and intra-part co-occurrence information of different body parts through a Graph Convolutional Network (GCN)\cite{kipf_2017_semi_gcn}. To the best of our knowledge, the proposed approach is the first to capture both inter and intra-part co-occurrence information through a GCN to address the problem of occluded pedestrian detection.
	
	\section{Method}
	As discussed above, occlusion is one of the most challenging problems in pedestrian detection. In case of occlusion, pedestrian body parts are either partially or fully occluded. Based on the observation that a human body generally has a fixed empirical ratio with limited flexibility of deformation, we propose a two-stage approach  that utilizes 
	spatial co-occurrence of different pedestrian body parts as a useful cue for heavily occluded pedestrian detection. 
	
	Fig.~\ref{fig:overall_network} shows the overall architecture of our proposed PSC-Net.~It consists of a standard pedestrian detection (PD) branch (Sec.~\ref{sec:spd}) and a part spatial co-occurrence (PSC) module (Sec.~\ref{sec:part_cooccurance}).~The standard pedestrian detection (PD) branch is based on Faster R-CNN~\cite{fasterrcnn_2015_nips} typically employed in existing pedestrian detection works~\cite{citypersons_2017_zhang,MGAN_2019_ICCV}. The part spatial co-occurrence (PSC) module encodes both inter and intra-part co-occurrence information of different body parts. The PSC module 
	comprises of two components. In the first component, intra-part co-occurrence of a pedestrian body part is captured by utilizing the corresponding RoI feature. As a result, an enhanced part feature representation is obtained. This representation is used as an input to the second component for capturing the inter-part co-occurrence of spatially adjacent body parts, leading to a final enhanced feature representation that encodes both intra and inter-part information. This final enhanced feature representation of a candidate proposal is then deployed as an input to the later part of the pedestrian detection (PD) branch which performs final bounding-box regression and classification. 
	
	Next, we describe the standard pedestrian detection (PD) branch, followed by a detailed presentation of our part co-occurrence (PSC) module (Sec.~\ref{sec:part_cooccurance}).
	
	\subsection{Standard Pedestrian Detector} \label{sec:spd}
	Here, we describe the standard pedestrian detection (PD) branch that is based on the popular Faster R-CNN framework~\cite{fasterrcnn_2015_nips} and typically employed in several pedestrian detection methods~\cite{citypersons_2017_zhang,MGAN_2019_ICCV}.
	
	The PD branch consists of a  backbone network, a region proposal network (RPN), region-of-interest (RoI) pooling layer and a classification network for final bounding-box regression and classification. In PD branch, the image is first feed into the backbone network and the RPN generates a set of candidate proposals for the  input image. For each candidate proposal, a fixed-sized feature representation is obtained through RoI pooling. Finally, this fixed-sized feature representation is passed through a classification network that outputs the classification score and the regressed bounding box locations for the corresponding proposal. The loss function ${L}_{f}$ of the standard pedestrian detection (PD) branch is given as follows:
	\begin{equation}
	{L}_{f} = {L}_{rpn\_cls}+{L}_{rpn\_reg} + {L}_{rcnn\_cls}+{L}_{rcnn\_reg},
	\label{loss:faster_rcnn}
	\end{equation}
	where ${L}_{rpn\_cls}$ and ${L}_{rpn\_reg}$ are the classification loss and bounding box regression loss of RPN, respectively, and ${L}_{rcnn\_cls}$ and ${L}_{rcnn\_reg}$ are the classification and bounding box regression loss of the classification network. In PD branch, Cross-Entropy loss is used as classification loss, and Smooth-L1  loss as bounding-box regression loss.
	
	\noindent \textbf{Limitations:} To handle heavy occlusions, several recent two-stage pedestrian detection approaches~\cite{ShanshanCVPR18, Zhang_2018_ECCV, MGAN_2019_ICCV} extend the PD branch by exploiting additional visible bounding-box (VBB) annotations along with the standard full body information. However, this reliance on additional VBB information implies that two sets of annotations are required for pedestrian detection training.  Further, these approaches obtain a fixed-sized proposal representation by performing a pooling operation (\emph{e.g.}, RoI Pool or RoiAlign) on the high-level features from the later layer of the backbone network (\emph{e.g.}, $conv5$ of VGG). Alternatively, several anchor-free methods~\cite{Weiliu_2019_CVPR, Song_2018_ECCV} have investigated the effective utilization of features from multiple ($conv$) layers of the backbone network for pedestrian detection. 
	
	In this work, we propose a two-stage pedestrian detection method, PSC-Net, to address heavy occlusions. 
	Our main contribution is the introduction of a \emph{part spatial co-occurrence} (PSC) module that only requires standard full body supervision and explicitly captures both inter and intra-part spatial co-occurrence information of different body parts.~Next, we describe the details of our PSC module. 
	
	\begin{figure}[t!]
		\centering
		\resizebox{\linewidth}{!}{
			\begin{tabular}{cc}
				\includegraphics[height=3.5cm]{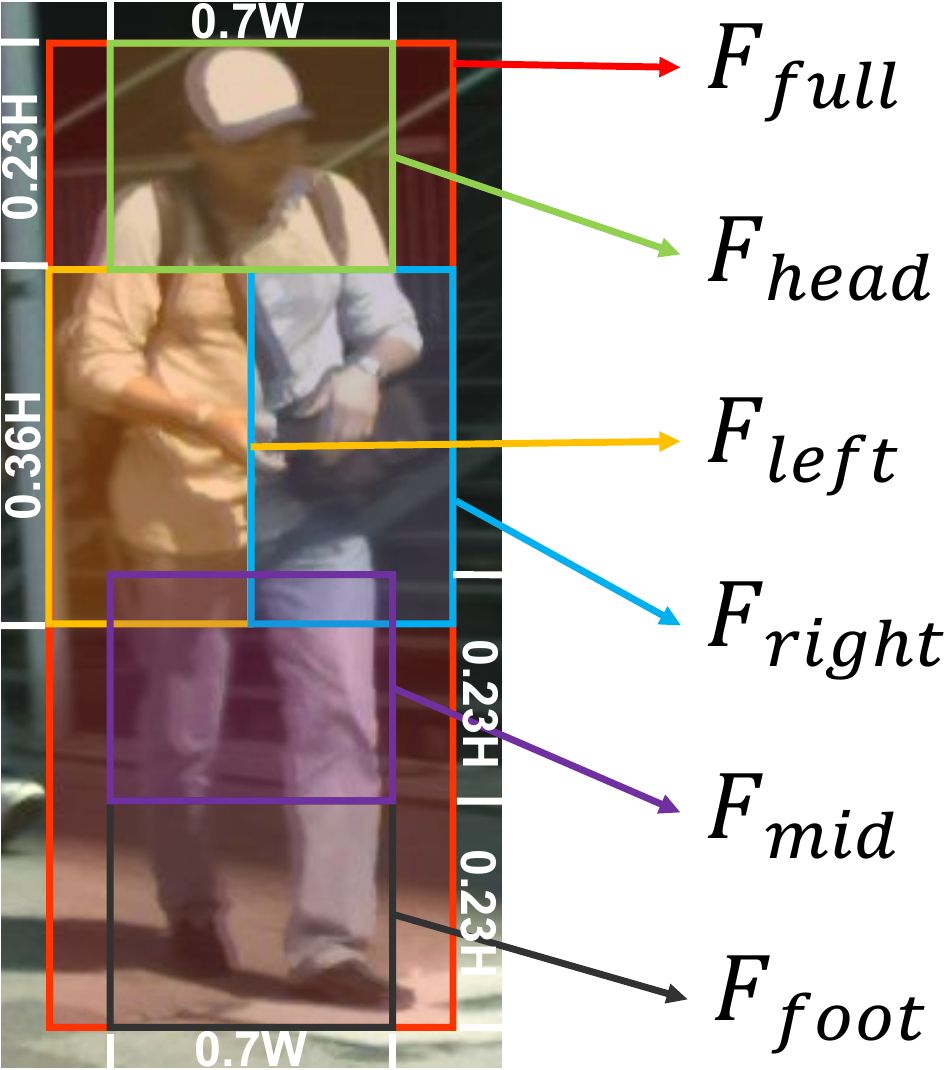}
				&\includegraphics[height=3.5cm]{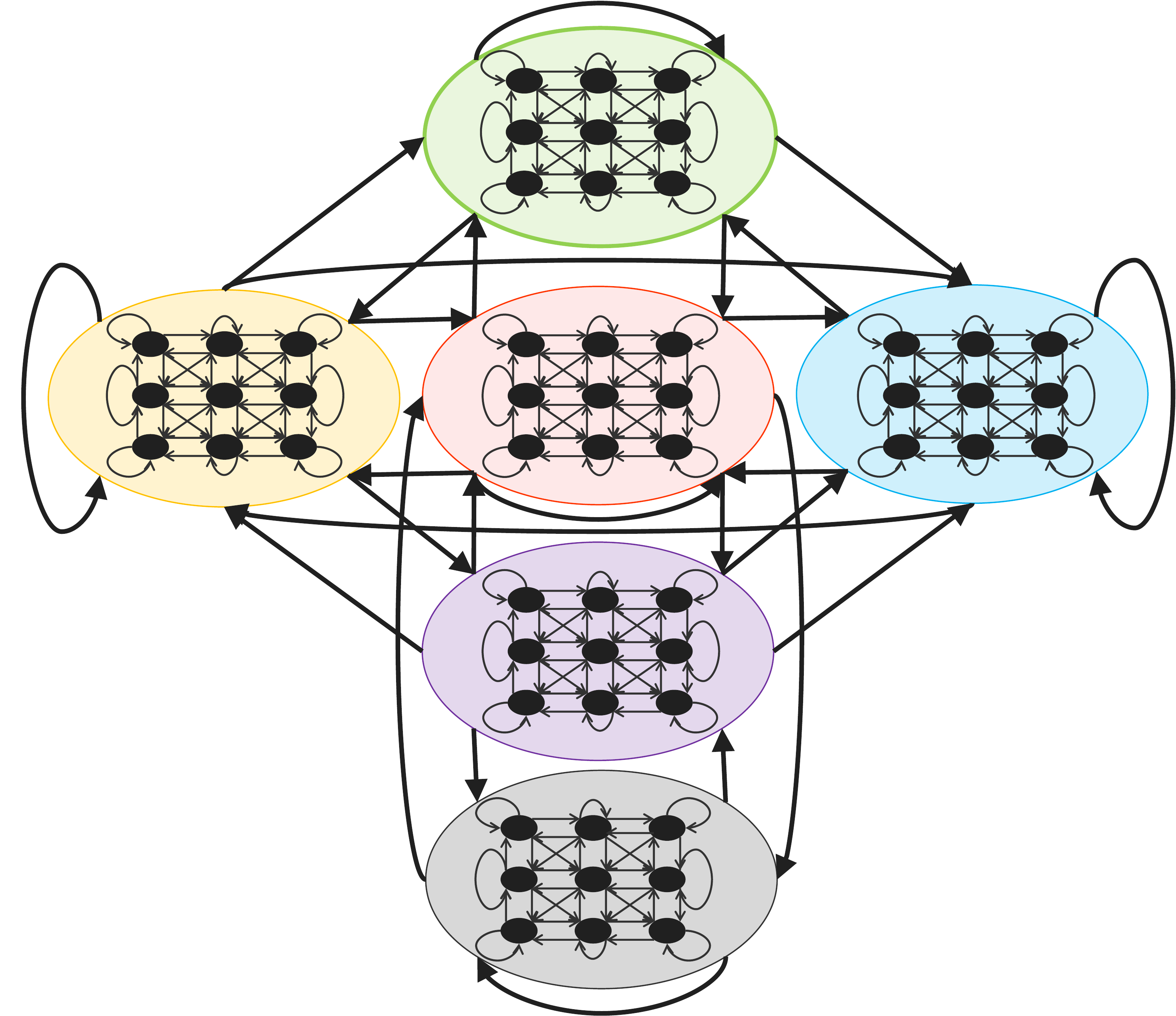}\\
				(a)\quad\quad\quad\quad\quad&(b)\\
		\end{tabular}}\\
		\caption{ (a) Full body pedestrian bounding-box partitioned into five parts based on empirically fixed ratio of human body. Each body part is shown with a different color and full body is in red. (b) Illustration of intra and inter-part spatial adjacency, used within our PSC module, to capture the spatial co-occurrence information. Our intra-part co-occurrence component employs a graph convolution layer to capture the spatial relation between different sub-regions of each body part. Differently, our inter-part component captures co-occurrence of spatially adjacent body parts using an additional graph convolution layer. Note that node colors in (b) are identical to the corresponding body parts in (a).}  
		\label{fig:prn}
	\end{figure}
	
	\subsection{Part Spatial Co-occurrence Module}
	\label{sec:part_cooccurance}
	In pedestrian detection, the task is to accurately localize the \emph{full body} of a pedestrian. This task is relatively easier in case of standard non-occluded pedestrians. However, it becomes particularly challenging in case of partial or severe occlusions. Here, we introduce a part spatial co-occurrence (PSC) module that utilizes spatial co-occurrence of different body parts captured through a Graph Convolutional Network (GCN)~\cite{kipf_2017_semi_gcn}. 
	In PSC module, the GCN is employed to capture intra-part and inter-part spatial co-occurrence by exploiting the topological structure of pedestrian. The intra-part co-occurrence is expected to improve the feature representation in scenarios where a particular body part is partially occluded whereas the inter-part co-occurrence targets severe occlusion of a particular body part. 
	
	Our PSC module neither requires pedestrian body part annotations nor rely on the use of an external pre-trained part model. Instead, it divides the full body bounding-box of a pedestrian into five parts ($ F_{head}$, $F_{left}$, $F_{right}$, $F_{mid}$, $F_{foot}$),  based on empirical fixed ratio of human body (see Fig.~\ref{fig:prn}), as in \cite{Zhang_2018_ECCV}. 
	The RoI pooling operation is performed on each body part (five) as well as the full body (${F}_{D}$), resulting in six RoI pooled features for each proposal. 
	
	As described above, the RoI pooling operation is performed for five body parts as well as the full body, resulting in an increased feature dimension. Therefore, a direct utilization of all these RoI features will drastically increase the computational complexity of our PSC module. Note that the Faster R-CNN and its pedestrian detection adaptations \cite{citypersons_2017_zhang, MGAN_2019_ICCV, Zhang_2018_ECCV} commonly use a single RoI pooling only on the $conv5$ features of VGG, resulting in $512$ channels. To maintain a similar number of channels as in Faster R-CNN and its pedestrian detection adaptations~\cite{ShanshanCVPR18, Zhang_2018_ECCV, MGAN_2019_ICCV}, we introduce an additional $1\times1$ convolution in RoI pooling strategy that significantly reduces the number of channels (572 in total). Consequently, RoI pooled features of each body part and the full body has only $64$ and $256$ channels, respectively.
	
	\subsubsection{Intra-part Co-occurrence}
	\label{sec:Intra_part_Co_occurrence}
	Here, we enhance RoI pooled feature representation of each body part by considering their intra-part co-occurrence. For instance, consider a scenario where head part $F_{head}$ is partially occluded, thereby making top-part of the head invisible.~Our intra-part co-occurrence component aims to capture spatial relation between different sub-regions (\emph{e.g.} eyes and ears) within a RoI feature $F_m$ of a body part (\emph{e.g.}, $F_{head}$) through graph convolutional layer, 
	
	\begin{equation}
	\tilde{F}_{m} = \sigma(A_{s}F_m W_s) \label{eq:eq2}
	\end{equation}
	
	\begin{figure}[t!]
		\includegraphics[width=\linewidth]{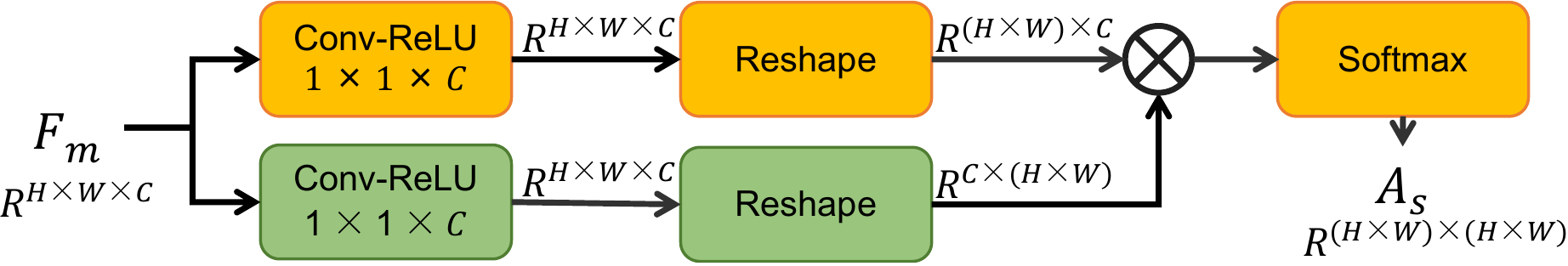}
		\caption{Computation of intra-part spatial adjacency  matrix $\mathcal{A}_{s}$. Each body part RoI feature $F_m$ is first passed through two parallel convolutional layers that are cascaded by ReLU.~The resulting features are re-shaped to perform matrix multiplication, followed by a softmax operation.} 
		\label{fig:matrix_fig}
	\end{figure}
	
	where $\sigma$ is the ReLU activation, $W_s \in R^{C \times C}$ is a learnable parameter matrix, $C$ is the number of channels (64 for each body part and 256 for full body) and 
	$A_{s}\in R^{H\times W\times H \times W}$ is intra-part spatial adjacency  matrix of a graph $\mathcal{G}_s=({\mathcal{V}_s,\mathcal{A}_s})$. Here, each pixel within the RoI region is treated as a graph node. In total, there are  $H\times W$ nodes  $\mathcal{V}_s$ in the graph. 
	
	The intra-part spatial adjacency matrix $A_{s}$ is computed as follow, and is also shown in Fig.~\ref{fig:matrix_fig}. We first pass the  RoI feature $F_m \in R^{H \times W\times C}$ through two parallel  $1 \times 1$  convolution layers that are cascaded by ReLU activation. The resulting outputs are re-shaped  prior to performing a matrix multiplication, followed by  a softmax operation to compute the intra-part spatial adjacency  matrix $A_{s}$. 
	
	The  output from the graph convolution layer (Eq.~\ref{eq:eq2}) is $\tilde{F}_{m} \in R^{H \times W\times C}$. This output $\tilde{F}_{m}$ is first added to its input $F_m$ (original RoI feature),  followed by passing it through a fully connected layer to obtain a $d$ dimensional  enhanced part feature. The enhanced part features $F_e$  of all six parts (five parts and full body) are further used to capture  inter-part co-occurrence described next.
	
	\subsubsection{Inter-part Co-occurrence}
	\label{sec:Inter_part_Co_occurrence}
	Our inter-part co-occurrence component is designed to improve feature representation, especially in case of severe  occlusion of a particular body part.  
	
The traditional convolution layer only captures information from a small spatial neighborhood, defined by the kernel size (\emph{e.g.} $3\times3$), and is therefore often ineffective to encode inter-part co-occurrence information of a pedestrian.
To address this issue, we introduce an additional graph convolutional layer in our PSC module, that captures  inter-part relationship of different body parts and the full body of a pedestrian. We treat each  part (including full body) as separate nodes ($\mathcal{V}_p$) of a graph $\mathcal{G}_p = (\mathcal{V}_p,\mathcal{A}_p)$,  where $\mathcal{A}_p$ denote the spatial adjacency matrix capturing the neighborhood relationship of different nodes. The graph convolutional operation is used to improve the node features $F_e$ as follows, 
\begin{equation}
{\tilde{F_e}}= \sigma((I - \mathcal{A}_p) F_e W_p) 
\end{equation}

where $F_e \in R^{n \times d}$ is enhanced features of $n$ parts, $A_p$ is the spatial adjacency matrix,  $W_p \in R^{d \times d}$ is learnable parameter matrix and $\sigma$ is ReLU activation. Here,  $n=6$ is the number of nodes, and $d=1024$ is channels of each node. $(I - \mathcal{A}_p)$ is used to conduct Laplacian smoothing \cite{2018_AAAI_Laplacian_smoothing} to propagate the node features over the graph.
		
The spatial adjacency matrix $\mathcal{A}_p$ captures the relation between different graph nodes.
During heavy occlusion, enhanced part features of a particular node may not contain relevant body part information. Therefore, it is desired to assign lesser weights to the edges linking such nodes in $\mathcal{A}_p$. 
To this end, we introduce a self-attention scheme to each edge that assigns learnable weight $a_{ij}$.  
The input of the self attention is the concatenated features of node $i$ and node $j$.
The self attention of each edge $a_{ij}$ is computed by a fully-connected operations followed by a sigmoid activation. 
The unnormalized spatial adjacency matrix $\hat{\mathcal{A}}_p(i, j)$ is defined as:
\begin{equation}
\hat{\mathcal{A}}_p(i, j) = \left\{\begin{matrix}
a_{ij} &   \text{part}\ i \ \text{and part} \ j \ \text{are spatial adjacent}\\
0& \text{otherwise}
\end{matrix}\right.
\end{equation}
The spatial adjacency matrix ${\mathcal{A}}_p(i, j)$ is computed by conducting $L2$ normalize in each row of $\hat{\mathcal{A}}_p(i, j)$.

Afterwards, we employ full connected layer to merge all the features ${\tilde{F_e}} \in R^{n \times d}$ into a $d$-dimensional feature vector. The resulting \emph{enriched} features are then utilized as an input to the classification network which predicts the final classification score and bounding-box regression.

	\section{Experiments}
	\noindent \textbf{Datasets:} We perform experiments on two datasets: CityPersons~\cite{citypersons_2017_zhang} and Caltech~\cite{Dollar_2012_PAMI}. CityPersons~\cite{citypersons_2017_zhang} consists of 2975 training, 500 validation, and 1525 test images. Caltech~\cite{Dollar_2012_PAMI} contains 11 sets of videos. First 6 sets (0-5) are used for training, the last 5 sets (6-10) are used for testing. To get a large amount of training data, we sample the videos with 10Hz. The training sets consists of 42782 images in total.
	
	\noindent \textbf{Evaluation Metrics:} We report the performance using log-average miss rate (MR) throughout our experiments. Here, MR is computed over the false positive per image (FPPI) range of $[10^{-2}, 10^0]$ \cite{Dollar_2012_PAMI}. On CityPersons, we report the results across two different occlusion degrees: Reasonable (\textbf{R}) and Heavy Occlusion (\textbf{HO}). For both \textbf{R} and \textbf{HO} sets, the height of pedestrians is larger than 50 pixels. The visibility ratio in \textbf{R} set is larger than $65\%$ whereas in \textbf{HO} it ranges from $20\%$ to $65\%$. In addition to these sets, the results are reported on combined (\textbf{R + HO}) set on Caltech. 
	
	\noindent \textbf{Implementation Details:} For both datasets, we train our network on a NVIDIA GPU with the mini-batch consisting of two images per GPU. Adam \cite{Adam} solver is selected as optimizer. In case of CityPersons, we fine-tune pre-trained ImageNet VGG model~\cite{simonyan2014vgg} on the trainset of the CityPersons. We follow the same experimental protocol as in~\cite{citypersons_2017_zhang} and employ two fully connected layers with 1024 instead of 4096 output dimensions. The initial learning rate is set to $10^-4$ for the first 8 epochs, and decay it by a factor of 10 for another 3 epochs. For Caltech, we start with a model that is pre-trained on CityPersons. An initial learning rate of $10^-5$  is used for first 3 training epochs and decay it to $10^-6$ for another 1 training epoch. Both the source code and models will be publicly available. 
	
	\subsection{CityPersons Dataset}
	\begin{table}[t!]
	\centering

	\resizebox{1.0\linewidth}{!}{
	\begin{tabular}{cp{3cm}p{3cm}|c|cc}
		\hline
		Baseline (PD)          &Intra-Part Co (Sec.~\ref{sec:Intra_part_Co_occurrence})        &Inter-Part Co  (Sec.~\ref{sec:Inter_part_Co_occurrence})           &\textbf{R} & \textbf{HO} \\
		\hline
		\checkmark   &   &             &13.8 & 56.8      \\
		\checkmark   &\multicolumn{1}{c}{\checkmark}   &{}   &11.8 &53.1      \\
		\checkmark   &{}   &    \multicolumn{1}{c|}{\checkmark}                     &12.5 & 52.1           \\
		\checkmark   &\multicolumn{1}{c}{\checkmark}   &\multicolumn{1}{c|}{\checkmark}        & \textbf{10.6} & \textbf{50.2}           \\
		\hline
	\end{tabular}}
	\caption{Impact of integrating our intra-part (Intra-Part Co) and inter-part (Inter-Part Co) co-occurrence into the baseline on CityPersons val. set. Performance is reported in terms of log-average miss rate and best results are boldfaced. Our final PSC-Net that integrates both the intra-part and inter-part co-occurrence achieves consistent improvement in performance, with gains of 3.2\% and 6.6\% on the \textbf{R} and \textbf{HO} sets, respectively, over the baseline.}
	\label{ablation_combine_impact}
\end{table}

\noindent\textbf{Baseline Comparison:}
\begin{table}[t]
	\centering

	\resizebox{1.0\linewidth}{!}{ 
	\begin{tabular}{c|c|c|c|c}
		\hline
		Methods                 & Data (Visibility) & Input Scales &  \textbf{R}   &  \textbf{HO}  \\ \hline
		TLL \cite{Song_2018_ECCV}        &        -         &     $1\times$     &     14.4      &     52.0      \\ \hline
		F.RCNN+ATT-part \cite{ShanshanCVPR18}  &    $\geq65\%$    &     $1\times$     &     16.0      &     56.7      \\
		F.RCNN+ATT-vbb \cite{ShanshanCVPR18}   &                  &     $1\times$     &     16.4      &     57.3      \\
		Repulsion Loss \cite{XinlongWangCVPR18} &                  &     $1\times$     &     13.2      &     56.9      \\
		Adaptive-NMS \cite{Songtao_2019_CVPR}  &                  &     $1\times$     &     11.9      &     55.2      \\
		MGAN \cite{MGAN_2019_ICCV}        &                  &     $1\times$     &     11.5      &     51.7      \\
		\textbf{PSC-Net (Ours)}                  &                  &     $1\times$    & \textbf{10.6} & \textbf{50.2} \\ \hline						            
		
		OR-CNN \cite{Zhang_2018_ECCV}      &    $\geq50\%$    &     $1\times$    &     12.8      &     55.7      \\
		MGAN \cite{MGAN_2019_ICCV}        &                  &     $1\times$    &     10.8      &     46.7      \\
		\textbf{PSC-Net (Ours)}                   &                  &     $1\times$    & \textbf{10.3} & \textbf{44.9} \\ \hline
		ALFNet \cite{WeiLiuECCV18}        &    $\geq0\%$     &     $1\times$     &     12.0      &     52.0      \\
		CSP \cite{Weiliu_2019_CVPR}       &                  &     $1\times$     &     11.0      &     49.3      \\
		MGAN \cite{MGAN_2019_ICCV}        &                  &     $1\times$     &     11.3      &     42.0      \\
		\textbf{PSC-Net (Ours)}                  &                  &     $1\times$     & \textbf{10.5} & \textbf{39.5} \\ \hline
		Repulsion Loss \cite{XinlongWangCVPR18} &    $\geq65\%$    &     $1.3\times$     &     11.5      &     55.3      \\
		Adaptive-NMS \cite{Songtao_2019_CVPR}  &                  &     $1.3\times$     &     10.8      &     54.0      \\
		MGAN \cite{MGAN_2019_ICCV}        &                  &     $1.3\times$    &     10.3      &     49.6      \\
		\textbf{PSC-Net (Ours)}                  &                  &     $1.3\times$     & \textbf{9.8}  & \textbf{48.3} \\ \hline
		OR-CNN \cite{Zhang_2018_ECCV}      &    $\geq50\%$    &     $1.3\times$     &     11.0      &     51.3      \\
		MGAN \cite{MGAN_2019_ICCV}        &                  &     $1.3\times$     &      9.9      &     45.4      \\
		\textbf{PSC-Net (Ours)}                   &                  &     $1.3\times$     & \textbf{9.6}  & \textbf{43.6} \\ \hline
		Bi-box \cite{Chunluan_2018_ECCV}       &    $\geq30\%$    &     $1.3\times$     &     11.2      &     44.2      \\
		FRCN +A +DT \cite{Chunluan_2019_ICCV}  &                  &     $1.3\times$     &     11.1      &     44.3      \\
		MGAN \cite{MGAN_2019_ICCV}        &                  &     $1.3\times$     &     10.5      &     39.4      \\
		\textbf{PSC-Net (Ours)}                   &                  &     $1.3\times$     & \textbf{9.9}  & \textbf{37.2} \\ \hline
	\end{tabular}}
	\caption{Comparison with the state-of-the-art (in terms of log-average miss rate) on the CityPersons validation set. In each case, best results are boldfaced. Our PSC-Net achieves superior performance on both datasets, compared to existing methods. When using same input scale (1.3x), training data visibility ($\geq30\%$) and backbone (VGG), PSC-Net provides absolute gains of 7.1\% and 2.2\% over FRCN +A +DT~\cite{Chunluan_2019_ICCV} and MGAN~\cite{MGAN_2019_ICCV}, respectively on the \textbf{HO} set. } 
	\label{tab:sota_city_val}
\end{table}
As discussed earlier, our design is the introduction of a part spatial co-occurrence (PSC) module that explicitly captures both intra-part (Sec.~\ref{sec:Intra_part_Co_occurrence}) and inter-part (Sec.~\ref{sec:Inter_part_Co_occurrence})  co-occurrence information of different body parts. Tab.~\ref{ablation_combine_impact} shows the impact of integrating our intra-part and inter-part co-occurrence components in the baseline. 
For fair comparison, all results in Tab.~\ref{ablation_combine_impact} are reported by using the same set of ground-truth pedestrian examples during training. All ground-truth pedestrian examples which are at least 50 pixels tall with visibility $\geq65$ are utilized for training. Further, the input scale of $1.0\times$ is employed during the experiments.
The integration of each component into the baseline results in consistent improvement in performance. Further, our final PSC-Net that integrates both the intra-part and inter-part co-occurrence achieves absolute gains of 3.2\% and 6.6\% on \textbf{R} and \textbf{HO} sets, respectively, over the baseline. These results demonstrate that both components are required to obtain optimal performance. 

\begin{table}[t!]

	\centering
	\begin{tabular}{c|c|c}
		\hline
		Method                       &   \textbf{R}   &  \textbf{HO}   \\ \hline
		Adaptive Faster RCNN \cite{citypersons_2017_zhang} &     13.0      &     50.5      \\
		MS-CNN \cite{cai_2019_cascade}                     &     13.3      &     51.9 \\
		Rep. Loss \cite{XinlongWangCVPR18}         &     11.5      &     52.6      \\
		OR-CNN \cite{Zhang_2018_ECCV}            &     11.3      &     51.4      \\
		Cascade MS-CNN  \cite{cai_2019_cascade}           &     11.6      &     47.1 \\               
		Adaptive-NMS \cite{Songtao_2019_CVPR}    &     11.4      &       -   \\
		MGAN \cite{MGAN_2019_ICCV}               &     \textbf{9.3}       &     41.0       \\
		\textbf{PSC-Net (Ours)}             &     \textbf{9.3}      & \textbf{37.0} \\  \hline
	\end{tabular}
	\caption{State-of-the-art comparison (in terms of log-average miss rate) on CityPersons test set. The test set is withheld and results are obtained by sending our PSC-Net detection predictions for evaluation to the authors of CityPersons~\cite{citypersons_2017_zhang}. Our PSC-Net outperforms existing methods on both \textbf{R} and \textbf{HO} sets. On heavy occluded \textbf{HO} set, PSC-Net achieves an absolute gain of 4.0\% over the state-of-the-art~\cite{MGAN_2019_ICCV}. In each case, best results are boldfaced.}
	\label{tab:citypersons_test_results}
\end{table}

\noindent\textbf{State-of-the-art Comparison:} Here, we perform a comparison of our PSC-Net with state-of-the-art pedestrian detection methods in literature. Tab.~\ref{tab:sota_city_val} shows the comparison on CityPersons validation set. Note that existing approaches utilize different set of ground-truth pedestrian examples for training. For fair comparison, we therefore select same set of ground-truth pedestrian examples (denoted as data (visibility) in Tab.~\ref{tab:sota_city_val}) and input scale, when performing a comparison with each state-of-the-art
method. Our PSC-Net achieves superior performance on all these settings for both \textbf{R} and \textbf{HO} sets, compared to the state-of-the-art methods.

When using an input scale of $1\times$ and data visibility ($\geq65\%$), the attention-based approaches, F.RCNN+ ATT-part \cite{ShanshanCVPR18} and F.RCNN+ATT-vbb \cite{ShanshanCVPR18}, obtain log-average miss rates of 16.0, 56.7 and 16.4, 57.3 on the \textbf{R} and \textbf{HO} sets. The work of~\cite{XinlongWangCVPR18} based on Repulsion Loss obtains log-average miss rates of 13.2 and 56.9 on the \textbf{R} and \textbf{HO} sets, respectively. The Adaptive-NMS approach~\cite{Songtao_2019_CVPR} that applies a dynamic suppression threshold and learns density scores obtains log-average miss rates of 11.9 and 55.2 on the \textbf{R} and \textbf{HO} sets, respectively. MGAN~\cite{MGAN_2019_ICCV} learns a spatial attention mask using VBB information to modulate full body features and achieves log-average miss rates of 11.5 and 51.7 on the \textbf{R} and \textbf{HO} sets, respectively. Our PSC-Net outperforms MGAN, without using VBB supervision, on both sets with log-average miss rates of 10.6 and 50.2. When using same data visibility but $1.3\times$ input scale, Adaptive-NMS~\cite{Songtao_2019_CVPR} and MGAN~\cite{MGAN_2019_ICCV} achieve log-average miss rates of 54.0 and 49.6, respectively on the \textbf{HO} set. The same two approaches report 10.8 and 10.3 on the \textbf{R} set. 
PSC-Net achieves superior results with log-average miss rates of 9.8 and 48.3 on the \textbf{R} and \textbf{HO} sets, respectively. On this dataset, the best existing results of 10.5 and 39.4 are reported~\cite{MGAN_2019_ICCV} on \textbf{R} and \textbf{HO} sets, respectively when using an input scale of $1.3\times$ and data visibility ($\geq30\%$). Our PSC-Net outperforms the state-of-the-art~\cite{MGAN_2019_ICCV} with log-average miss rates of 9.9 and 37.2 on the two sets. 

Tab.~\ref{tab:citypersons_test_results} shows the comparison on CityPersons test set. Among existing methods, the multi-stage Cascade MS-CNN  \cite{cai_2019_cascade} consisting of a sequence of detectors trained with increasing IoU thresholds  obtains log average miss rate of 47.1 on the \textbf{HO} set. MGAN~\cite{MGAN_2019_ICCV} obtains a log average miss rate of 41.0 on the same set. Our PSC-Net significantly reduces the error by 4.0\% over MGAN on the \textbf{HO} set. Similarly, PSC-Net also improves the performance on the \textbf{R} set. 

\begin{table}[t]
	\centering

	\begin{tabular}{c|c|c|c|c}
		\hline
		Detector                &   Occl.    &  \textbf{R}  &  \textbf{HO}  & \textbf{R+HO} \\ \hline\hline
		CompACT-Deep \cite{Cai_2015_ICCV}   &  $\times$  &     11.8     &     65.8      &     24.6      \\
		ATT-vbb \cite{ShanshanCVPR18}     & \checkmark &     10.3     &     45.2      &     18.2      \\
		MS-CNN \cite{ZhaoweiCaiECCV16}     &  $\times$  &     10.0     &     59.9      &     21.5      \\
		RPN+BF \cite{zhang2016faster}     &  $\times$  &     9.6      &     74.4      &     24.0      \\
		SA-F.RCNN \cite{li2018scale}      &  $\times$  &     9.7      &     64.4      &     21.9      \\
		SDS-RCNN \cite{GarrickBrazilICCV17}  &  $\times$  &     7.4      &     58.6      &     19.7      \\
		F.RCNN \cite{citypersons_2017_zhang}  &  $\times$  &     9.2      &     57.6      &     20.0      \\
		GDFL \cite{Lin_2018_ECCV}       &  $\times$  &     7.9      &     43.2      &     15.6      \\
		Bi-Box \cite{Chunluan_2018_ECCV}      & \checkmark &     7.6      &     44.4      &     16.1      \\
		AR-Ped \cite{Brazil_2019_CVPR}     &  $\times$  & 6.5 &     48.8      &     16.1      \\
		FRCN +A +DT \cite{Chunluan_2019_ICCV} & \checkmark &     8.0      &     37.9      &       -       \\
		MGAN \cite{MGAN_2019_ICCV}       & \checkmark &     6.8      &     38.2      &     13.8      \\ \hline
		\textbf{PSC-Net (Ours)}               & \checkmark &     \textbf{6.4}      & \textbf{34.8} & \textbf{12.7} \\ \hline
	\end{tabular}
	\caption{State-of-the-art comparison (in terms of log-average miss rate) on Caltech test set. Best results are boldfaced in each case. Our PSC-Net provides consistent improvements (over) on all sets. On the \textbf{HO} set, PSC-Net outperforms the best reported results~\cite{Chunluan_2019_ICCV} by reducing the error from 37.9 to 34.8.}
	\label{tab:caltech_results}
\end{table}

\begin{figure}[t!]
	\centering
	
	\resizebox{\linewidth}{!}{
	\begin{tabular}{c}
		\includegraphics[height=2.0cm]{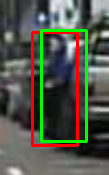}
		\includegraphics[height=2.0cm]{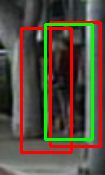}  
		\includegraphics[height=2.0cm]{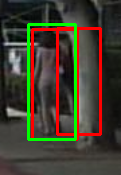}  
		\includegraphics[height=2.0cm]{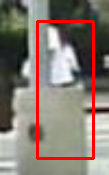}  
		\includegraphics[height=2.0cm]{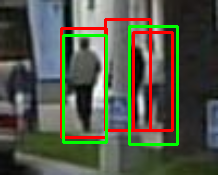}
		\\           
		(a) AR-Ped \cite{Brazil_2019_CVPR}\\
		\includegraphics[height=2.0cm]{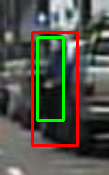}    
		\includegraphics[height=2.0cm]{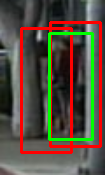}
		\includegraphics[height=2.0cm]{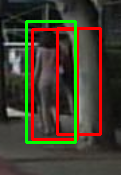}  
		\includegraphics[height=2.0cm]{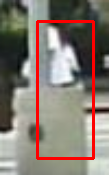}
		\includegraphics[height=2.0cm]{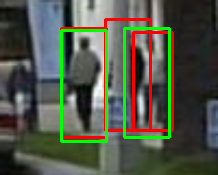}      
		\\   
		(b) MGAN \cite{MGAN_2019_ICCV}\\
		\includegraphics[height=2.0cm]{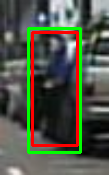}
		\includegraphics[height=2.0cm]{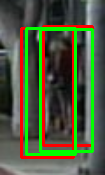}  
		\includegraphics[height=2.0cm]{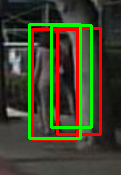}  
		\includegraphics[height=2.0cm]{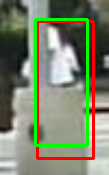}
		\includegraphics[height=2.0cm]{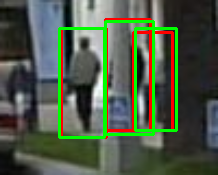}   \\
		(c) PSC-Net\\
	\end{tabular}}
	\caption{Qualitative detection comparison of (c) PSC-Net with (a) AR-Ped \cite{Brazil_2019_CVPR} and (b) MGAN \cite{MGAN_2019_ICCV} under occlusions on caltech test images. Here, all detection results are obtained using the same false positive per image criterion. The red boxes denote the ground-truth whereas the detector predictions are in green. For better visualization, the detected regions are cropped from the corresponding images.} 
	\label{fig:my_label_caltech_qual}
\end{figure}

\subsection{Caltech Dataset} 
Finally, we evaluate our PSC-Net on Caltech and compare it with state-of-the-art approaches in literature.
Tab.~\ref{tab:sota_city_val} shows the comparison on Caltech test set under three sets: \textbf{R}, \textbf{H} and \textbf{R + HO}. Among existing methods, ATT-vbb \cite{ShanshanCVPR18}, Bi-box \cite{Chunluan_2018_ECCV}, FRCN +A +DT \cite{Chunluan_2019_ICCV} and MGAN \cite{MGAN_2019_ICCV}, address the problem of occlusions by utilizing VBB information. On the \textbf{R}, \textbf{H} and \textbf{R + HO} subsets, AR-Ped \cite{Brazil_2019_CVPR}, FRCN +A +DT \cite{Chunluan_2019_ICCV} and MGAN~\cite{MGAN_2019_ICCV} report best existing performance, respectively. PSC-Net achieves superior detection performance on all three subsets with log average miss rates of 6.4, 34.8 and 12.7, respectively.  
Fig.~\ref{fig:my_label_caltech_qual} displays example detections showing a visual comparison of our PSC-Net with recently introduced AR-Ped \cite{Brazil_2019_CVPR} and MGAN \cite{MGAN_2019_ICCV} under occlusions. All detection results in Fig.~\ref{fig:my_label_caltech_qual} are obtained using same false positive per image (FPPI) criterion. Our PSC-Net provides improved detections under these occlusion scenarios compared to the other two approaches.

	\section{Conclusion}
	We proposed a two-stage approach, PSC-Net, for occluded pedestrian detection. Our PSC-Net consists of a standard pedestrian detection branch and a part spatial co-occurrence (PSC) module. The focus of our design is the  PSC module that is designed to capture intra-part and inter-part spatial co-occurrence of different body parts through a Graph Convolutional Network (GCN). Our PSC module only requires standard full body supervision and exploits the topological structure of pedestrian. Experiments are performed on two popular datasets: CityPersons and Caltech. Our results clearly demonstrate that the proposed PSC-Net significantly outperforms the baseline in all cases. Further, the PSC-Net sets a new state-of-the-art on both datasets.

	\ifCLASSOPTIONcaptionsoff
	\newpage
	\fi

	

	\bibliographystyle{plain}
	\bibliography{pscnet}

	

\end{document}